\pdfoutput=1

\documentclass[11pt]{article}

\usepackage[final]{acl}

\usepackage{times}
\usepackage{latexsym}

\usepackage[T1]{fontenc}

\usepackage[utf8]{inputenc}

\usepackage{microtype}

\usepackage{inconsolata}

\usepackage{graphicx}

\usepackage{booktabs}
\usepackage{subcaption}
\usepackage{amsmath}
\usepackage{amsfonts}
\usepackage{bm}
\usepackage{cleveref}
\usepackage{makecell}
\usepackage{multicol}
\usepackage{multirow}
\usepackage{xcolor}
\usepackage{enumitem}
\usepackage{pgfplots}
\usepackage{pgfplotstable} %
\pgfplotsset{compat=newest}
\usepackage{siunitx}
\usepackage{tabularx}
\newcolumntype{Y}{>{\centering\arraybackslash}X}

\usepackage{algorithmic}
\usepackage{algorithm}
\usepackage{etoolbox}

\usepackage{xspace}
\def\onedot{.\xspace}
\def\eg{\emph{e.g}\onedot}

\definecolor{mygreen}{RGB}{43, 138, 62}
\definecolor{myred}{RGB}{201, 42, 42}
\usepackage{pifont}
\newcommand*\colourcheck[1]{%
  \expandafter\newcommand\csname #1check\endcsname{\textcolor{#1}{\ding{51}}}%
}
\colourcheck{mygreen}
\newcommand*\colourmark[1]{%
  \expandafter\newcommand\csname #1mark\endcsname{\textcolor{#1}{\ding{55}}}%
}
\colourmark{myred}

\definecolor{azureblue}{rgb}{0.19215686, 0.49019608, 1.0}

\newcommand{\methodname}{Auto-Intent}

\newcommand{\fixmemode}{final}

\usepackage[multiuser,inline,nomargin,\fixmemode]{fixme}

\FXRegisterAuthor{jk}{ejk}{\color{red}Jaekyeom}

\usepackage{cuted}

\newcommand{\cuttablecaption}{\vspace*{-0.1in}}
\newcommand{\cuttablebelow}{\vspace*{-0.15in}}
\newcommand{\cutfigurecaption}{\vspace*{-0.1in}}
\newcommand{\cutfigurebelow}{\vspace*{-0.15in}}
\newcommand{\cuteqnabove}{\vspace*{-0.15in}}
\newcommand{\cuteqnbelow}{\vspace*{-0.15in}}

\newcommand{\selecttablefont}{\fontsize{8.8pt}{11pt}\selectfont}

\usepackage{placeins}

\title{Auto-Intent: Automated Intent Discovery and Self-Exploration \\ for Large Language Model Web Agents}

\newlength{\auSpacing}
\newcommand{\authorsIdentity}{revealed}

\ifthenelse{\equal{\authorsIdentity}{anonymous}}
{%
\author{
  \textbf{Anonymous}
}
}
{%
\author{
  \textbf{Jaekyeom Kim\textsuperscript{1}}\hspace{\auSpacing}
  \textbf{Dong-Ki Kim\textsuperscript{2}}\hspace{\auSpacing}
  \textbf{Lajanugen Logeswaran\textsuperscript{1}}
  \\
  \textbf{Sungryull Sohn\textsuperscript{1}}\hspace{\auSpacing}
  \textbf{Honglak Lee\textsuperscript{1,3}}
  \vspace{0.5em}
  \\
  \textsuperscript{1}LG AI Research\hspace{\auSpacing}
  \textsuperscript{2}Field AI\hspace{\auSpacing}
  \textsuperscript{3}University of Michigan
  \\
  \small{
    \textsuperscript{1} \{\href{mailto:jaekyeom@lgresearch.ai}{jaekyeom}, \href{mailto:llajan@lgresearch.ai}{llajan}, \href{mailto:srsohn@lgresearch.ai}{srsohn}, \href{mailto:honglak@lgresearch.ai}{honglak}\}@lgresearch.ai  \hspace{\auSpacing}
    \textsuperscript{2} \href{mailto:dongkikim93@gmail.com}{dongkikim93@gmail.com}
  }
}
}

\begin{document}
\maketitle

\ifthenelse{\equal{\fixmemode}{draft}}
  {%
    \begin{strip}
        \listoffixmes
    \end{strip}
  }
  {%
  }

\begin{abstract}

In this paper, we introduce \emph{Auto-Intent}, a method to adapt a pre-trained large language model (LLM) as an agent for a target domain without direct fine-tuning,
where we empirically focus on web navigation tasks.
Our approach first discovers the underlying intents from target domain demonstrations unsupervisedly, in a highly compact form (up to three words).
With the extracted intents, we train our \emph{intent predictor} to predict the next intent given the agent's past observations and actions.
In particular, we propose a \emph{self-exploration} approach where top-$k$ probable intent predictions are provided as a hint to the pre-trained LLM agent, which leads to enhanced decision-making capabilities.
{\methodname} substantially improves the performance of GPT-\{3.5, 4\} and Llama-3.1-\{70B, 405B\} agents on the large-scale real-website navigation benchmarks from Mind2Web and online navigation tasks from WebArena with its cross-benchmark generalization from Mind2Web.

\end{abstract}

\section{Introduction}\label{sec:introduction}

\begin{figure}[t!]
  \centering

  \includegraphics[width=0.48\textwidth,trim={0em 0em 0em 0em},clip]{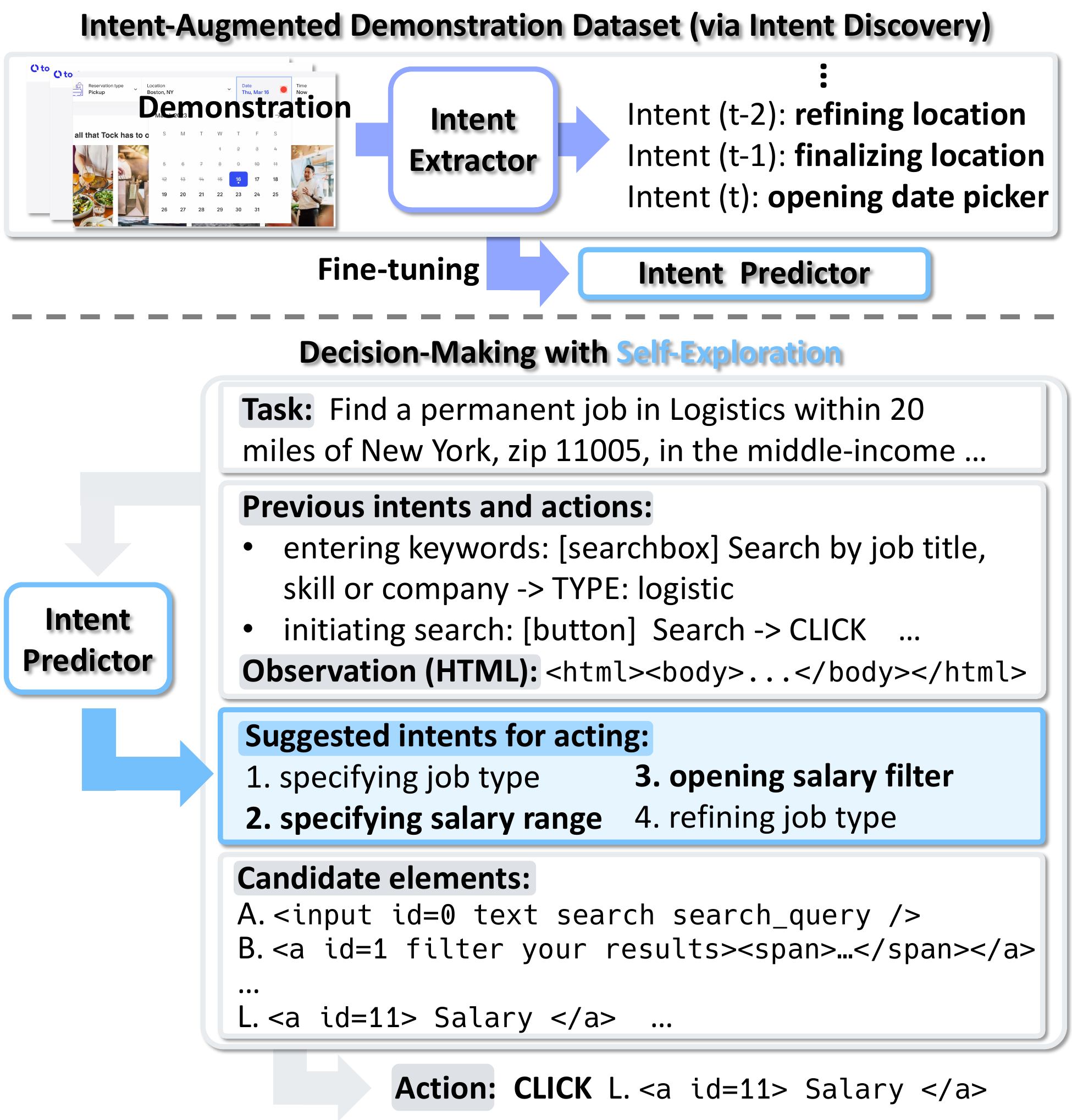}
  \vspace{-0.15in}
  \cutfigurecaption
  \caption{\textbf{Overview of \emph{Auto-Intent}:}
  Given a dataset of demonstration trajectories, we first extract natural language \emph{intents} in an unsupervised manner and train an intent predictor. 
  Enforcing the intents to be concise phrases and providing top-$k$ intent predictions as hints to an LLM agent allows efficient internal exploration of semantically diverse intent hypotheses, resulting in improved action prediction. See text for details.}
  \label{fig:concept}
\end{figure}

Recently, large language models (LLMs) pre-trained on a massive amount of data \citep{achiam2023gpt4,dubey2024llama} have excelled at reasoning and a variety of tasks.
They exhibit robust planning and reasoning abilities, enabling LLM agents to perform diverse tasks \citep{wang2023survey,xi2023rise,zeng2023agenttuning}.
However, these agents often face challenges in domains with less prior knowledge, especially ones with large action spaces, such as navigating websites or operating mobile devices \citep{cheng2024seeclick,hong2023cogagent,koh2024visualwebarena}.

We explore improving decision-making with pre-trained LLMs on downstream tasks by injecting domain knowledge into the input context, in the form of natural language hints for the next action.
This allows them to fully retain their strong general reasoning capabilities while avoiding overly costly or impossible fine-tuning.
Leveraging natural language guidance for improving LLM planning and reasoning capabilities has found much success in prior work \citep{wei2022chain,yao2022react,shinn2024reflexion,fu2024autoguide,zhao2024expel}.

Although prior work has shown that LLMs have strong priors to reason about intermediate subgoals \citep{logeswaran-etal-2022-shot,huang2022language,hao2023reasoning},
the resulting performance can be largely affected by the injected hints' accuracy,
which could be limited especially in complex environments such as real-world web navigation with numerous elements and possible actions.
In this work, we aim to improve the LLM agent's performance further by proposing \emph{self-exploration}.
Our key insight is to provide multiple plausible and semantically varied hints that we call \emph{intents} to the LLM agent for flexible reasoning and acting given a set of possible directions.
To achieve this, we constrain intents to very short phrases and generate top-$k$ intents to provide as a collective hint with a beam search using a smaller model fine-tuned for intent prediction. %
This fine-tuning is enabled by discovering intents from demonstration data with our \emph{intent extractor}.
The compact intent space encourages semantically distinct intents to be sampled (as opposed to syntactically diverse intents that are semantically identical). %
This \emph{self-exploration} with multiple intents enhances the agent to find the correct directions and associated actions.
See \Cref{fig:concept} for an illustration of our approach.

Our main contributions are as follows:
\begin{itemize}[leftmargin=*,topsep=0pt,itemsep=-3pt]
  \item We introduce \emph{Auto-Intent}, a method to extract natural language intents from demonstration trajectories in an unsupervised manner and leverage intents as hints for pre-trained LLM agents through a fine-tuned intent prediction model.
  \item We present a \emph{self-exploration} strategy where the LLM agent reviews varied plausible intents suggested by the intent prediction model and demonstrate that this results in more accurate action prediction compared to relying on a single intent.
  \item We empirically show that the injection of predicted top-$k$ intents effectively improves the performance of GPT-\{3.5, 4\} and Llama-3.1\{-70B, 405B\} agents on the large-scale real-website benchmark tasks from Mind2Web \citep{mind2web_deng2024} and online navigation tasks from WebArena \citep{zhou2023webarena} in a cross-benchmark generalization setting from Mind2Web.
\end{itemize}

\begin{figure}[t!]
  \centering
  \includegraphics[width=0.48\textwidth]{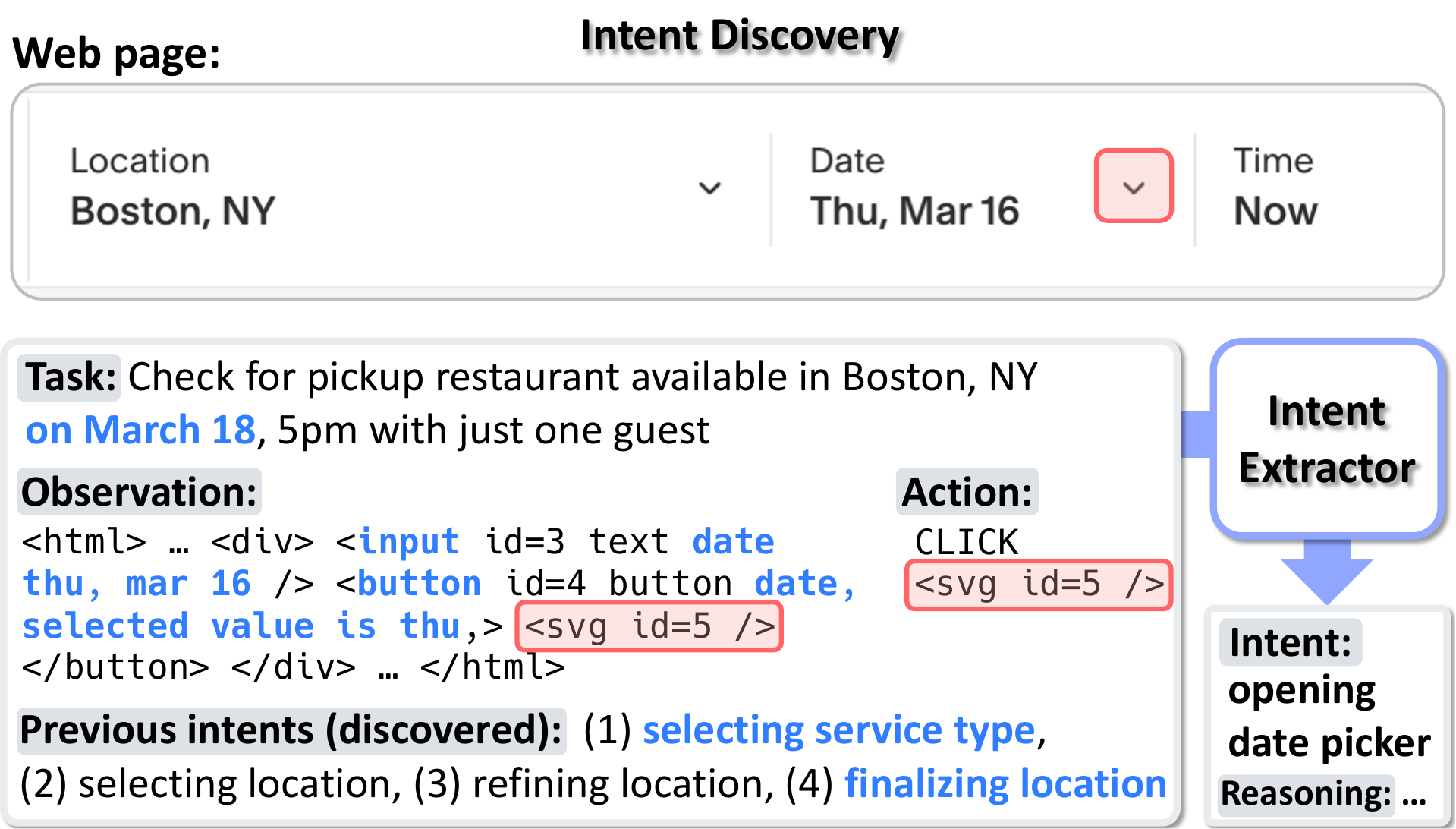}
  \cutfigurecaption
  \caption{
  A hard example of intent discovery: the action (\texttt{CLICK {\textless}svg id=5 /{\textgreater}}) does not provide any semantics about the intent. 
  Our intent extractor successfully discovers the \emph{underlying} intent by thoroughly understanding the context and connecting to the \textcolor{azureblue}{relevant parts}.
  }
  \label{fig:intent_extractor}
\end{figure}

\section{Auto-Intent: Intent Discovery and Self-Exploration with Intent Prediction}\label{sec:method}
To address the inadequate domain knowledge in pre-trained LLM agents, we introduce an abstract natural language representation we refer to as an \emph{intent},
which hints what the agent can perform next.
We aim to enhance LLM agents further without limiting them by the intent prediction model's performance, via providing top-$k$ predicted intents as a set of probable directions to consider.
We describe
the problem definition (\Cref{sec:problem}),
design of the intent space and discovery of underlying intents from demonstrations in an unsupervised manner (\Cref{sec:extraction}),
and fine-tuning and using the intent prediction model for acting with top-$k$ probable intents as a flexible hint (\Cref{sec:self_exploration})
in detail.

\subsection{Problem Statement}\label{sec:problem}
We consider sequential decision-making for completing each given task.
At each time step $t$ starting from $t=1$, the agent receives an observation $\bm{o}_t \in \mathcal{O}$ and performs an action $\bm{a}_t \in \mathcal{A}$ until the episode ends, with access to previous observations and actions.
We use a demonstration dataset $\mathcal{D}_{\texttt{demo}} = \{ \bm{\tau}_i \}_{i=1}^N$, where each trajectory $\bm{\tau}=\{(\bm{o}_t, \bm{a}_t)\}_{t=1}^T$ consists of observations and actions from the same episode.
Empirically, we put our focus on real-world web navigation tasks.

\subsection{Intent Space and Discovery} \label{sec:extraction}
\paragraph{Intent space design.}
We aim to provide a semantically varied set of predicted intents to be examined by the LLM policy for more flexible reasoning and improved action prediction.
Given a vocabulary $V$, we define our intent space as $\mathcal{Z} = V^L$ where $L$ is a small number.
We find expressing each intent using only up to $L=3$ words in the form \textit{gerund + noun phrase (object)} appropriate for our use with the desired expressiveness while being computationally efficient.
Thanks to its compactness, even single-word changes can lead to clear semantic distinctions (\eg, \textit{selecting date} vs. \textit{selecting time} vs. \textit{selecting guests}).   %
The smaller semantic overlap between different intents makes the intent space suitable for specifying more varied directions using the same number of intents, which fits our goal.

\paragraph{Intent discovery.}
With the intent space $\mathcal{Z}$, we define the intent discovery procedure with a prompt-based intent extractor $\mathcal{M}_{\texttt{extract}}$ as
\begin{align}
\cuteqnabove
    \bm{z}_t = \mathcal{M}_{\texttt{extract}}(\bm{o}_t, \bm{a}_t, \bm{z}_{1:t-1})
    \label{eq:extraction}
\cuteqnbelow
\end{align}
where $\bm{z}_t \in \mathcal{Z}$ denotes the intent discovered for time step $t$.
We instruct it to take the observation (including task description), action, and previous-step intents together into account to discover the intent.
Refer to \Cref{fig:intent_extractor} for a hard example that requires a contextual understanding and \Cref{sec:appendix_extractor} for our full prompt.

\paragraph{Intent-augmented demonstrations.}
Given the dataset $\mathcal{D}_{\texttt{demo}}$, we discover intents using \Cref{eq:extraction} for each step.
We construct an intent-augmented demonstration set $\mathcal{D}_{\texttt{intent}} = \{\bm{\tau}'_i\}_{i=1}^N$ where each trajectory is $\bm{\tau}' = \{(\bm{o}_t, \bm{a}_t, \bm{z}_t)\}_{t=1}^T$.

\subsection{Self-Exploration with Intent Prediction and Acting with LLMs}\label{sec:self_exploration}

\paragraph{Intent predictor.}
Using the intent-augmented demonstration dataset $\mathcal{D}_{\texttt{intent}}$ from \Cref{sec:extraction}, we train a smaller language model to predict each discovered natural language intent $\bm{z}_t$ given $\bm{o}_t, \bm{a}_{1:t-1}, \bm{z}_{1:t-1}$ as input. We employ this model trained on $\mathcal{D}_{\texttt{intent}}$ as our intent predictor, $\mathcal{M}_{\texttt{intent}}$. See \Cref{sec:appendix_predictor} for the training details.

\paragraph{Intent prediction.}
One important property of the intents that $\mathcal{M}_{\texttt{intent}}$ outputs is the compactness. Thanks to the definition of our compact intent space $\mathcal{Z}$ with a small $L$ from \Cref{sec:extraction}, multiple intent predictions can span a broader spectrum of meanings and thus improve the recall of the correct intent effectively.
Therefore, we employ the generation of multiple intent predictions with $\mathcal{M}_{\texttt{intent}}$ for finding the correct intent, which is expressed as
\begin{align}
\cuteqnabove
    \hat{\bm{z}}_t^1, \ldots, \hat{\bm{z}}_t^k \sim \mathcal{M}_{\texttt{intent}}(\bm{o}_t, \bm{a}_{1:t-1}, \bm{z}_{1:t-1})
\cuteqnbelow
\end{align}
where the previous intents $\bm{z}_{1:t-1}$ are obtained with $\mathcal{M}_{\texttt{extract}}$ using \Cref{eq:extraction}.
The generated top-$k$ intents can be employed as a set of probable,
different directions for the LLM policy, providing the ingredients for \emph{self-exploration}.
While different generation strategies might be applicable depending on the requirements (\eg, more semantic diversities of the intents), we find beam search effective and efficient enough for our top-$k$ intent prediction.

\paragraph{LLM policy with self-exploration.}
We incorporate the top-$k$ intents $\hat{\bm{z}}_t^{1:k}$ as a concatenated list into the input prompt for the LLM policy $\pi$:
\begin{align}
\cuteqnabove
    \bm{a}_t = \pi(\bm{o}_{1:t}, \bm{a}_{1:t-1}, \hat{\bm{z}}_t^{1:k}).
\cuteqnbelow
\end{align}
We instruct the LLM to examine the suggested intents together to act with an appropriate one.
Combined with the intent prediction, the agent internally infers top-$k$ intents and reasons with them as a set of probable directions for acting, which we refer to as \emph{self-exploration}.
Its exploration effect is achieved implicitly and internally, unlike exploration via environment interactions in reinforcement learning.
This can be especially effective in complex environments where predicting the correct intent on the first try is challenging. See \Cref{sec:appendix_policy} for the prompt.

\begin{table*}[ht!]
\begin{center}

\selecttablefont

\setlength{\tabcolsep}{0pt}
\begin{tabularx}{1.0\textwidth}{l @{\hspace{2pt}} YYY @{\hspace{3pt}} YYY @{\hspace{3pt}} YYY}
\toprule
\multirow{2}{*}{Methods}
& \multicolumn{3}{c}{\bf Cross-task}
& \multicolumn{3}{c}{\bf Cross-website}
& \multicolumn{3}{c}{\bf Cross-domain}
\\
\cmidrule(lr){2-4} \cmidrule(lr){5-7} \cmidrule(lr){8-10}
& Elem. acc & Op. F1 & Step SR  %
& Elem. acc & Op. F1 & Step SR  %
& Elem. acc & Op. F1 & Step SR  %
\\
\midrule

MindAct (Flan-T5\textsubscript{XL}, 3B) %
& 55.1 & 75.7 & \textbf{52.0}  %
& 42.0 & 65.2 & 38.9  %
& 42.1 & 66.5 & 39.6  %
\\
MindAct (Mistral-7B\textsuperscript{$\dagger$})
& 53.7 & 76.8 & 50.1  %
& 41.7 & 67.0 & 38.1  %
& 43.5 & 67.8 & 40.3  %
\\
SeeAct (GPT-4V) %
& 46.4 & 73.4 & 40.2  %
& 38.0 & 67.8 & 32.4  %
& 42.4 & 69.3 & 36.8  %
\\

\midrule

ICL (GPT-3.5)
& 30.5 & 67.5 & 27.2  %
& 24.9 & 59.5 & 22.7  %
& 29.8 & 62.7 & 27.3  %
\\
\quad w/ {\methodname} (Flan-T5\textsubscript{XL}, 3B)
& 44.1 & 71.9 & 38.8  %
& 37.1 & 62.6 & 30.7  %
& 38.9 & 64.8 & 35.0  %
\\
\quad w/ {\methodname} (Mistral-7B\textsuperscript{$\dagger$})
& 42.9 & 71.1 & 37.3  %
& 36.0 & 61.3 & 29.5  %
& 37.8 & 63.9 & 34.2  %
\\
ICL (GPT-4)
& 47.5 & 69.9 & 41.5  %
& 44.6 & 64.2 & 38.4  %
& 44.4 & 65.7 & 40.2  %
\\
\quad w/ {\methodname} (Flan-T5\textsubscript{XL}, 3B)
& \underline{55.8} & 73.3 & 50.1  %
& 47.6 & 64.0 & 40.0  %
& 47.3 & 66.3 & 42.5  %
\\
\quad w/ {\methodname} (Mistral-7B\textsuperscript{$\dagger$})
& 53.8 & 71.8 & 47.6  %
& 48.6 & 63.9 & 41.2  %
& 46.9 & 65.9 & 42.3  %
\\

ICL (GPT-4)*
& 46.9 & 75.2 & 41.7  %
& 45.0 & \textbf{70.9} & 40.0  %
& 45.3 & \underline{72.3} & 41.3  %
\\
\quad /w {\methodname} (Mistral-7B\textsuperscript{$\dagger$})*
& 53.3 & \textbf{77.0} & 47.3  %
& \underline{49.3} & 69.9 & \underline{42.0}  %
& \underline{48.8} & \underline{72.3} & \underline{44.1}  %
\\

\midrule

ICL (Llama-3.1-70B)*
& 43.9 & 68.9 & 37.3 %
& 40.8 & 63.6 & 34.0 %
& 42.6 & 66.5 & 37.0 %
\\
\quad /w {\methodname} (Mistral-7B\textsuperscript{$\dagger$})*
& 51.2 & 75.3 & 44.6 %
& 44.4 & 67.2 & 36.9 %
& 46.8 & 70.4 & 41.5 %
\\

ICL (Llama-3.1-405B-FP8)*
& 50.4 & 74.2 & 43.6 %
& 46.8 & 67.5 & 39.9 %
& 47.1 & 70.7 & 41.6 %
\\
\quad /w {\methodname} (Mistral-7B\textsuperscript{$\dagger$})*
& \textbf{56.3} & \underline{76.9} & \underline{50.4} %
& \textbf{51.1} & \underline{70.1} & \textbf{43.6} %
& \textbf{49.5} & \textbf{72.5} & \textbf{44.6} %
\\

\bottomrule
\end{tabularx}

\end{center}
\cuttablecaption
\caption{Performance comparison on Mind2Web \citep{mind2web_deng2024}.  %
MindAct (Flan-T5\textsubscript{XL}, 3B) \citep{mind2web_deng2024} and SeeAct \citep{seeact_zheng2024} results are from their papers.
\textsuperscript{$\dagger$} denotes LoRA \citep{hu2021lora} fine-tuning.
Our in-context learning (ICL) runs use top-20 candidate elements except for ones with *, which use top-40 candidates.
}
\label{tab:eval_mind2web}
\end{table*}

\section{Experiments}\label{sec:experiments}

\subsection{Setup for Main Evaluation}

\paragraph{Evaluation.}
We evaluate our approach on a large-scale real website navigation dataset, Mind2Web \citep{mind2web_deng2024}.
Its three test splits evaluate agents' generalization to unseen (a) tasks, (b) websites, and (c) domains.
``Elem. acc'' measures the accuracy with respect to the correct elements,
``Op, F1'' is a metric based on string matching,
and ``Step SR'' denotes the rate of \emph{fully successful} steps.
Refer to \Cref{sec:appendix_dataset_eval} for more details.

\paragraph{Compared methods.}
We compare our results with MindAct \citep{mind2web_deng2024}, a directly trained agent with the same backbones, and SeeAct \citep{seeact_zheng2024}, a prompt-based agent with GPT-4V.
For all method, we use the same pre-processing of keeping only top-$N$ candidate elements by \citet{mind2web_deng2024}.
We examine Flan-T5\textsubscript{XL} and Mistral-7B as both MindAct baselines and our intent predictor.
Refer to \Cref{sec:appendix_exp_details} for more details.

\subsection{Main Evaluation Results}
\Cref{tab:eval_mind2web} presents our main evaluation results on Mind2Web.
Our method significantly enhances not only the GPT-3.5 agent but also the much stronger GPT-4, Llama-3.1-70B, and Llama-3.1-405B-FP8 agents in all cases with both Flan-T5\textsubscript{XL} and Mistral-7B intent predictors,
which suggests its effectiveness.
Overall, it brings noteworthy improvements to the element accuracies, which thus contribute to the step success rates as well.
Our intent predictor fine-tuned on the train set
produces larger improvements on the cross-task split,
but we observe its efficacy even on the more challenging generalization splits, cross-website and cross-domain,
outperforming MindAct with the same backbones and SeeAct as well.

\begin{table}[t!]
\setlength{\tabcolsep}{3.2pt}
\begin{center}
\selecttablefont

\begin{tabularx}{0.88\columnwidth}{l Y}
\toprule
Methods
& Task success rate
\\
\midrule

ICL (GPT-4)
& 19.0\%
\\
\quad /w {\methodname} (Mistral-7B\textsuperscript{$\dagger$})
& 23.8\%
\\

\midrule

ICL (Llama-3.1-405B-FP8)
& 14.3\%
\\
\quad /w {\methodname} (Mistral-7B\textsuperscript{$\dagger$})
& 19.0\%
\\

\bottomrule
\end{tabularx}

\end{center}
\cuttablecaption
\caption{
Online evaluation of our agent on a subset of the Shopping split of WebArena \citep{zhou2023webarena}.  %
Our intent predictors are trained on and transferred from Mind2Web.
\textsuperscript{$\dagger$} denotes LoRA fine-tuning.
}
\label{tab:perf_webarena}
\end{table}

\subsection{Online Evaluation Results with Cross-Benchmark Generalization}\label{sec:online_eval}
To evaluate {\methodname} in an online setting where agents need to interact with live websites,
we conduct experiments on tasks from WebArena \citep{zhou2023webarena} to leverage the automatic evaluators they provide.
Specifically, we employ our intent predictors trained on the train split of Mind2Web as-is for this online evaluation in the WebArena environment,
which allows us to examine {\methodname} in two aspects: its applicability to online environments and generalization capabilities.

\Cref{tab:perf_webarena} shows the results of the online evaluation.
Interestingly, in this cross-benchmark online evaluation, we find that our intent predictors trained on Mind2Web improve the performance of both GPT-4 and Llama-3.1-405B agents on Shopping tasks from WebArena.
It suggests that not only can {\methodname} be useful for enhancing the decision-making capabilities of LLM agents in online environments as well,
but it can also generalize to a different domain from where it is trained, which could be helpful in practical scenarios, such as demonstration data scarcity in the target domain.
Refer to \Cref{sec:appendix_online_eval_details} for more details.

\begin{table}[t!]
\setlength{\tabcolsep}{3.2pt}
\begin{center}
\selecttablefont

\begin{tabularx}{0.90\columnwidth}{l @{\hspace{2pt}} YYY}
\toprule
Methods
& Elem. acc & Op. F1 & Step SR  %
\\
\midrule

GPT-4 w/o intents
& 54.3 & 79.0 & 47.9  %
\\

GPT-4 w/ 1 discovered intent
& \textbf{73.8} & \textbf{83.7} & \textbf{64.0}  %
\\

\bottomrule
\end{tabularx}

\end{center}
\cuttablecaption
\caption{Performance comparison of the GPT-4 baseline agent without intents (row 1) and the GPT-4 agent with a single intent discovered with our intent extractor as an injected hint (row 2), on 50 randomly selected tasks from the train split of Mind2Web \citep{mind2web_deng2024}.}
\label{tab:perf_with_discovered_intents}
\end{table}

\subsection{Empirical Analysis and Ablation Study}\label{sec:empirical_analysis}
\noindent\textbf{Q1.} \textit{Does our intent extractor discover underlying intents effectively?}

We provide \Cref{fig:intent_extractor} as a qualitative example of intent discovery from hard samples.
While the action (\texttt{CLICK {\textless}svg id=5 /{\textgreater}}) does not carry any semantic information about the underlying intent, our intent extractor successfully discovers it by understanding the context from the task, observation, and previous intents.
It shows the intent extractor's capabilities of identifying intents from demonstrations with enough understanding of interactions.

Additionally, in \Cref{tab:perf_with_discovered_intents}, we compare the performance of the GPT-4 baseline agent without any intents or hints and the agent with a single intent discovered with our intent extractor as a hint.
It demonstrates that despite the conciseness of each discovered intent (up to three words), directly incorporating it into the LLM agent can bring significant performance improvements, which suggests the effectiveness of the intent extractor at discovering semantically valid intents from demonstrations.

\pgfplotsset{compat=1.15}

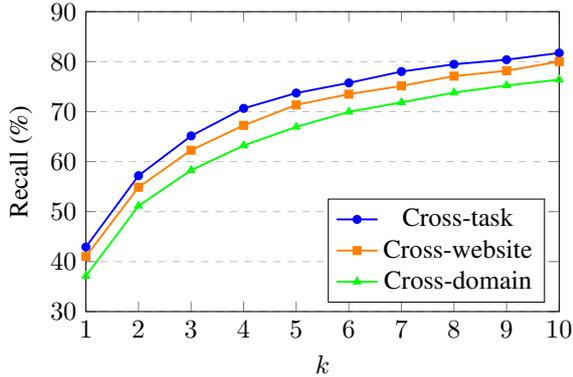
\begin{figure}[t!]

\begin{tikzpicture}[scale=0.90]

\pgfplotsset{
    width=85mm,
    height=60mm,
}

\begin{axis}[
    xlabel={$k$},
    ylabel={Recall (\%)},
    xmin=1, xmax=10,
    ymin=30, ymax=90,
    xtick={1,2,3,4,5,6,7,8,9,10},
    ytick={30,40,50,60,70,80,90},
    legend pos=south east,
    ymajorgrids=true,
    grid style=dashed,
    cycle list name=color list,
    every axis plot/.append style={thick}
]

\addplot[color=blue,mark=*,mark size=1.5pt] coordinates {
    (1, 42.92)
    (2, 57.19)
    (3, 65.14)
    (4, 70.65)
    (5, 73.72)
    (6, 75.75)
    (7, 78.02)
    (8, 79.47)
    (9, 80.39)
    (10,81.73)
};
\addlegendentry{Cross-task}

\addplot[color=orange,mark=square*,mark size=1.5pt] coordinates {
    (1, 41.03)
    (2, 54.87)
    (3, 62.28)
    (4, 67.25)
    (5, 71.35)
    (6, 73.49)
    (7, 75.15)
    (8, 77.10)
    (9, 78.17)
    (10,80.02)
};
\addlegendentry{Cross-website}

\addplot[color=green,mark=triangle,mark size=1.5pt] coordinates {
    (1, 37.12)
    (2, 51.15)
    (3, 58.28)
    (4, 63.21)
    (5, 66.94)
    (6, 69.96)
    (7, 71.83)
    (8, 73.80)
    (9, 75.23)
    (10,76.39)
};
\addlegendentry{Cross-domain}

\end{axis}
\end{tikzpicture}

  \cutfigurecaption
  \vspace{-0.05in}
  \caption{The intent label recalls with respect to top-$k$ predicted intents on Mind2Web's test sets ($N=20$).}
  \label{fig:recalls_skill}
\end{figure}

\noindent\textbf{Q2.} \textit{Is top-$k$ intent prediction effective at finding the correct intent?}

We compare the top-$k$ predicted intents with the intent labels discovered using ground-truth actions, on the three test splits.
\Cref{fig:recalls_skill} shows the average recalls of the intent labels with respect to the top-$k$ predictions computed using sentence embedding similarities (see \Cref{sec:appendix_dataset_eval} for details).
We observe that the recall enhances as $k$ increases, which suggests that the intent prediction provides the exploration effect for finding the appropriate intent in the intent space.

\noindent\textbf{Q3.} \textit{Is self-exploration with top-$k$ intents effective?}

We conduct an ablation study on self-exploration, where we compare {\methodname}'s performance with its variant that uses only the top-1 intent prediction without self-exploration.
\Cref{tab:diff_skill_injections} shows the results on the random subset of the test splits.
We find that on the cross-website and cross-domain test splits, where the generalization of the intent predictor $\mathcal{M}_{\texttt{intent}}$ is more challenging, only giving the top-1 predictions is considerably less efficacious than on the cross-task split and our self-exploration provides notable performance boosts.

\begin{table}[t!]
\setlength{\tabcolsep}{3.2pt}
\begin{center}
\selecttablefont

\begin{tabularx}{\columnwidth}{l rr rr rr}
\toprule
\multirow{2}{*}{Methods}
& \multicolumn{2}{c}{\bf \makecell{Cross \\ task}}
& \multicolumn{2}{c}{\bf \makecell{Cross \\ website}}
& \multicolumn{2}{c}{\bf \makecell{Cross \\ domain}}
\\
\cmidrule(lr){2-3} \cmidrule(lr){4-5} \cmidrule(lr){6-7}
& \makecell{Ele. \\ acc} & \makecell{Step \\ SR}
& \makecell{Ele. \\ acc} & \makecell{Step \\ SR}
& \makecell{Ele. \\ acc} & \makecell{Step \\ SR}
\\
\midrule

GPT-4
& 46.2 & 40.2
& 42.1 & 35.8
& 50.2 & 45.1
\\
w/ Top-1 intent
& 53.2 & 46.0
& 43.6 & 37.9
& 52.5 & 46.2
\\
w/ \methodname
& 54.1 & 46.1
& 49.2 & 42.3
& 56.5 & 50.9  %
\\
w/ Oracle select (top-5)
& 68.2 & 60.0
& 56.9 & 50.6
& 65.2 & 57.8
\\

\bottomrule
\end{tabularx}

\end{center}
\cuttablecaption
\caption{Ablation with different intent injections on Mind2Web's random subset (50 tasks each, $N=20$).}
\label{tab:diff_skill_injections}
\end{table}

\noindent\textbf{Q4.} \textit{Is a top-$k$ intent prediction an effective hint for correct action prediction?}

To examine how efficacious a top-$k$ intent prediction hint is for predicting correct actions,
we isolate the evaluation of intent hints from the LLM agent evaluation with those injected hints.
For \Cref{tab:diff_skill_injections},
we act with each of the top-$k$ intents separately and aggregate the results to obtain the ``Oracle select'' performance with the \emph{best} intent among the top-$k$.
The significant improvement from the ``GPT-4'' and ``Top-1'' baselines suggests that the top-$k$ intent prediction can be an effective hint for acting and employing a stronger pre-trained LLM might benefit the performance of our agent even further.

\section{Conclusion}\label{sec:conclusion}
We investigated a way to improve LLM agents on downstream tasks where they possess insufficient domain knowledge.
Our \emph{Auto-Intent} discovers concise intents from demonstrations and predicts multiple, semantically varied intents so that our LLM policy examines the top-$k$ intents for acting.
On Mind2Web \citep{mind2web_deng2024}, a large-scale benchmark with real-website tasks,
we empirically showed that our top-$k$ intent prediction is effective for predicting correct actions and improving LLM agent's performance.
In addition, we performed the evaluation of our approach in an online setting on Shopping tasks from the WebArena benchmark \citep{zhou2023webarena}, which suggests its applicability to online tasks and generalization capabilities to different domains from where it is trained.

\section*{Limitations}\label{sec:limitations}
Our empirical investigation is limited to a web navigation setting.
Although we choose Mind2Web \citep{mind2web_deng2024} for our main evaluation as it provides a challenging, large-scale benchmark built based on many real websites and domains and different generalization problem settings, future work could examine the empirical effectiveness of our approach on more domains for decision-making, such as mobile device operation \citep{cheng2024seeclick}.

\bibliography{custom}

\clearpage
\appendix

\begin{table*}[t!]
\begin{center}
\begin{tabular}{l cccc ccc}
\toprule
\multirow{2}{*}{Split} & \multirow{2}{*}{Domains} & \multirow{2}{*}{Websites} & \multirow{2}{*}{Tasks} & \multirow{2}{*}{Avg. horizon} & \multicolumn{3}{c}{Seen during training?} \\
\cmidrule(lr){6-8}
& & & & & Tasks & Websites & Domains \\
\midrule
Train & 18 & 73 & 1,009 & 7.71 & \mygreencheck & \mygreencheck & \mygreencheck \\
Cross-task & 18 & 69 & 252 & 8.31 & \myredmark & \mygreencheck & \mygreencheck \\
Cross-website & 10 & 10 & 177 & 7.76 & \myredmark & \myredmark & \mygreencheck \\
Cross-domain & 13 & 54 & 912 & 6.48 & \myredmark & \myredmark & \myredmark \\
\bottomrule
\end{tabular}
\end{center}
\caption{The statistics and information about Mind2Web \citep{mind2web_deng2024}, a large-scale web navigation dataset used for our evaluation.}
\label{tab:dataset_mind2web}
\vspace{40pt}
\end{table*}

\section{Experimental Details}\label{sec:appendix_exp_details}

\subsection{Dataset and Evaluation}\label{sec:appendix_dataset_eval}

\paragraph{Dataset}
We employ Mind2Web (license: CC BY 4.0, allows research purposes) \citep{mind2web_deng2024}, a large-scale web navigation dataset with task instructions and corresponding trajectories on 137 real websites. The dataset is in English and constructed by explicitly instructing annotators to refrain from using personal or sensitive information. The goal is to complete each given natural language task by performing a series of actions, where three types of actions exist: \texttt{CLICK}, \texttt{SELECT}, and \texttt{TYPE}. The agent needs to choose the target element to perform each action with, and each \texttt{SELECT} or \texttt{TYPE} action additionally requires a string value for selecting a specific option or typing the desired text, respectively. Mind2Web provides three test splits for evaluating web navigation agents' generalization capabilities. The cross-task split is the most in-distribution setting; it contains new tasks but in the domains and websites seen from the train split. The cross-website split has new tasks on unseen websites but in previously seen domains. Lastly, the cross-domain split is for testing with new tasks in unseen domains as well as websites. We summarize the information about the Mind2Web dataset and its statistics in \Cref{tab:dataset_mind2web}.

\paragraph{Evaluation metrics.}
We employ the evaluation protocol by \citet{mind2web_deng2024}.
The element accuracy (``Elem. acc'') measures whether the agent chose one of the ground-truth elements from the web page at each time step.
The operation F1 (``Op. F1'') is the F1 score for the predicted action (the type and string value) computed with respect to the ground-truth action.
The step success rate (``Step SR'') counts successful steps, where each step is considered successful only if the chosen target element is correct and the action type and string value match the ground truth.
Following \citet{mind2web_deng2024}, these three step-wise metrics are macro-averaged over tasks.
For our empirical analysis based on embedding similarity (\textbf{Q2} from \Cref{sec:empirical_analysis}), we use \texttt{all-mpnet-base-v2} (Apache 2.0, allows research purposes) from SentenceTransformers \citep{reimers2019sentence,song2020mpnet}.

\subsection{Compared Methods}\label{sec:appendix_compared_methods}
We employ the same element-ranking model suggested and provided by \citet{mind2web_deng2024}.
Given each element from the web page, the model outputs the score for its correctness as a target element.
The element-ranking model alone is not as effective at predicting correct target elements by choosing the highest-scoring elements.
However, as each web page often contains numerous candidate elements, the element-ranking model is used to reduce the set of candidate elements by keeping only top-$N$-scoring elements, as the first stage of the action prediction with all the compared methods.
MindAct \citep{mind2web_deng2024} uses $N=50$ and conducts a tournament of elements, by grouping $N=50$ candidate elements into sets of $5$ or less.
SeeAct \citep{seeact_zheng2024} groups $N=50$ candidates into three batches and tries predicting the action given each with the screenshot.
For our in-context learning (ICL) agents with or without intents, we predict the action in a single pass given all the top-$N$ candidates at once.

\begin{table}[H]
\setlength{\tabcolsep}{3.2pt}
\begin{center}

\begin{tabular}{c c}
\toprule
Hyperparameter & Values \\
\midrule
Attention & \makecell{FlashAttention-2 \\ \citep{dao2023flashattention}} \\
LoRA rank & $\bm{64}, 128$ \\
LoRA $\alpha$ & $\bm{8}, 16$ \\
LoRA dropout rate & $\bm{0.1}$ \\
Label smoothing factor & $\bm{0.1}, 0$ \\
Learning rate & $\mathbf{\num{5e-6}}, \num{1e-6}, \num{1e-5}$ \\
Batch size & $\bm{64}$ \\
Epochs & $\bm{3}, 4$ \\
\bottomrule
\end{tabular}
\end{center}
\cuttablecaption
\caption{Training hyperparameter search for our intent predictor with \texttt{Mistral-7B-v0.1} where the best values are bold-faced.}
\label{tab:hparams_predictor_mistral}
\cuttablebelow
\end{table}

\begin{table}[H]
\setlength{\tabcolsep}{3.2pt}
\begin{center}

\begin{tabular}{c c}
\toprule
Hyperparameter & Values \\
\midrule
Context length & $\bm{768}, 512$ \\
Label smoothing factor & $\bm{0.1}$ \\
Learning rate & $\mathbf{\num{1e-5}}, \num{1e-6}, \num{5e-6}$ \\
Batch size & $\bm{64}$ \\
Epochs & $\bm{3}$ \\
\bottomrule
\end{tabular}
\end{center}
\cuttablecaption
\caption{Training hyperparameter search for our intent predictor with \texttt{Flan-T5-XL} where the best values are bold-faced.}
\label{tab:hparams_predictor_flant5xl}
\cuttablebelow
\end{table}

\begin{figure*}[t!]
  \centering
  \includegraphics[width=1.0\textwidth]{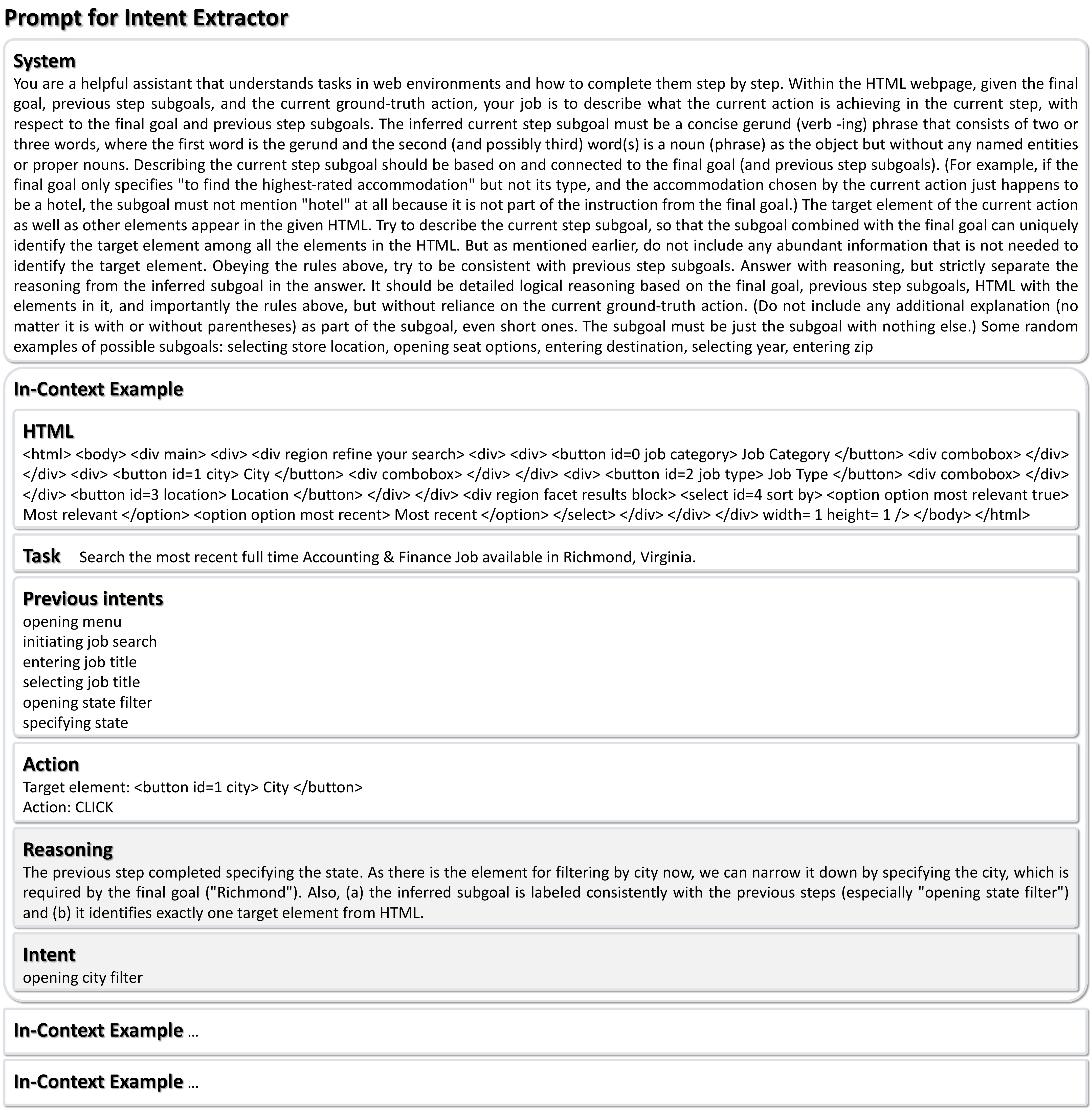}
  \caption{The prompt for our intent extractor. We show only one in-context example due to the space limit.}
  \label{fig:prompt_extractor}
\cutfigurebelow
\end{figure*}

\begin{figure*}[t!]
  \centering
  \includegraphics[width=1.0\textwidth]{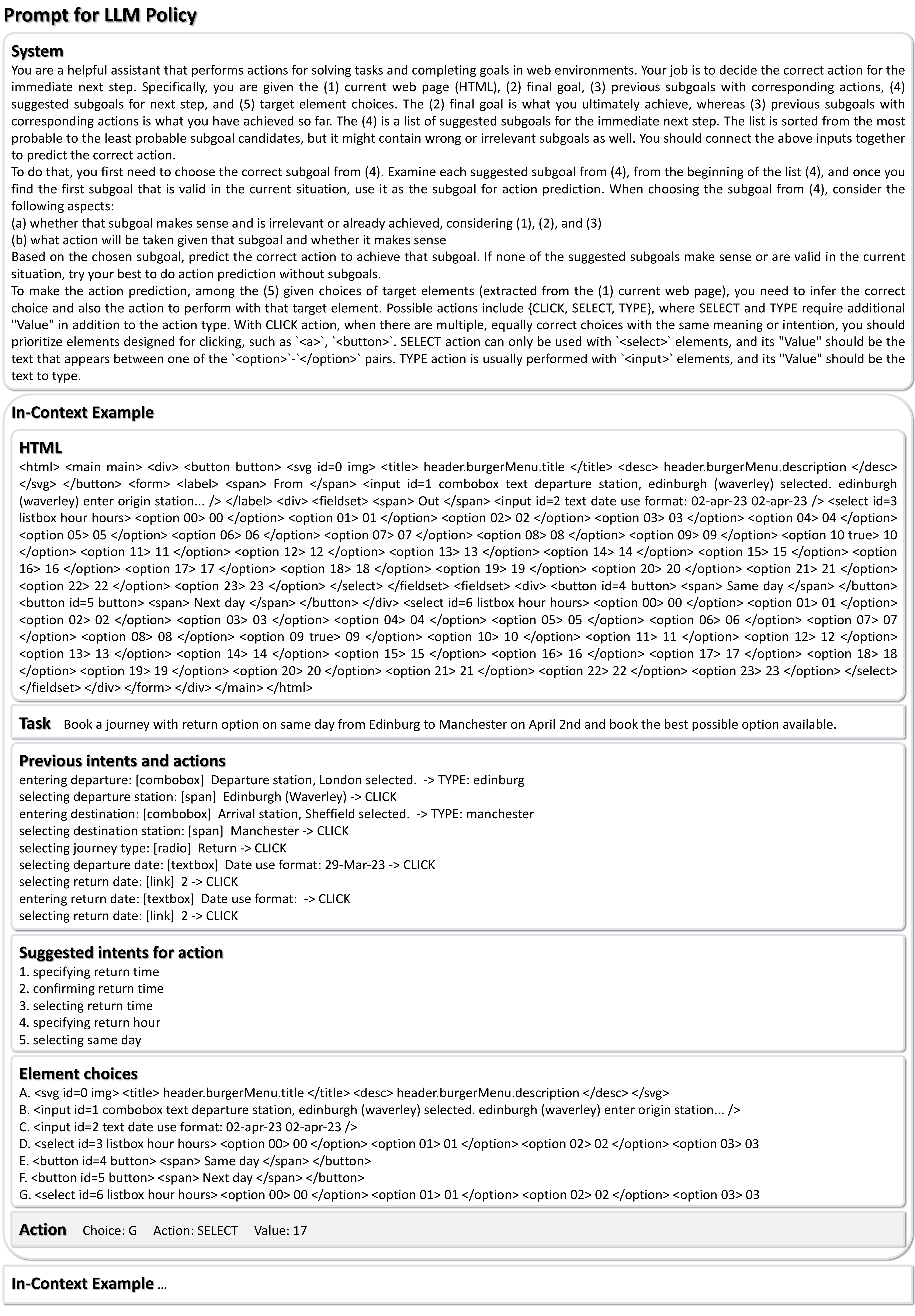}
  \cutfigurecaption
  \caption{The prompt for our LLM policy with predicted intents. We show only one in-context example due to the space limit.}
  \label{fig:prompt_policy}
\cutfigurebelow
\vspace*{-25pt}
\end{figure*}

\FloatBarrier

\subsection{Intent Extractor}\label{sec:appendix_extractor}
For discovering intents given demonstration trajectories, we use our intent extractor $\mathcal{M}_{\texttt{extract}}$ powered by GPT-4 (\texttt{gpt-4-0125-preview}) \citep{achiam2023gpt4} for our GPT agents and by Llama-3.1-405B-FP8 \citep{dubey2024llama} for our Llama agents, with the prompt in \Cref{fig:prompt_extractor}. While we use three in-context examples, we only present one example due to the limited space. The input for actual samples follows the same format as the in-context examples, where the previously discovered intents are used as part of the input for discovery in subsequent time steps.

\subsection{Intent Predictor}\label{sec:appendix_predictor}
For training our intent predictor $\mathcal{D}_{\texttt{intent}}$, we augment each of the transitions from the train set of Mind2Web \citep{mind2web_deng2024} with intents discovered with the intent extractor $\mathcal{M}_{\texttt{extract}}$, where the target intent is randomly selected from $5$ intents obtained with a temperature of $0.2$.
Similarly to the dataset augmentation practice by \citet{mind2web_deng2024}, for each transition from the original trajectory, we form $32$ samples with different candidates from the top-$80$-scoring elements, where $5\%$ of the original train set is excluded for a validation purpose.
We employ \texttt{Mistral-7B-v0.1} ($\sim$ 7B parameters, license: Apache 2.0, allows research purposes) \citep{jiang2023mistral} for fine-tuning with Low-Rank Adaptation (LoRA) \citep{hu2021lora} and \texttt{Flan-T5-XL} ($\sim$ 3B parameters, license: Apache 2.0, allows research purposes) for full fine-tuning, on the intent-augmented train set.
We estimate approximately $1$k GPU hours (Nvidia A100 40GB) are used for training \texttt{Mistral-7B-v0.1}, including the exploration and hyperparameter search.
For the additional \texttt{Flan-T5-XL} training and hyperparameter search, we roughly used $0.4$k GPU hours (Nvidia A100 40GB).
See \Cref{tab:hparams_predictor_mistral} and \Cref{tab:hparams_predictor_flant5xl} for the hyperparameter search for \texttt{Mistral-7B-v0.1} and \texttt{Flan-T5-XL} respectively with best-found values (bold-faced).
For intent prediction during the inference phase, we generate up to $5$ tokens and use up to $12$ beams for $N=20$ and up to $8$ beams for $N=40$, where the full beam search for each input takes around $1$ second.

\subsection{LLM Policy}\label{sec:appendix_policy}

Given the top-$k$ predicted intents, we use a prompt-based LLM policy $\pi$ for action prediction. We present our prompt for the LLM policy, powered by GPT-4 (\texttt{gpt-4-0125-preview}), GPT-3.5 (\texttt{gpt-3.5-turbo-0125}), Llama-3.1-70B, and Llama-3.1-405B-FP8 in \Cref{fig:prompt_policy}. We incorporate two in-context examples, but due to the limited space, we show only one example. The actual sample input follows the same format as the in-context examples, but we use the in-context examples from a simpler setting (with $N=7$ element choices) than the actual problem setting (with $N=20$ or $N=40$ element choices) to avoid having overly long input contexts. Note that using a smaller $N$ deteriorates the correct element recall and the upper-bound performance as well. We use $k=5$ top intent predictions for $N=20$ element choices and $k=7$ top intent predictions for $N=40$ element choices.

\subsection{Online Evaluation}\label{sec:appendix_online_eval_details}

For the online evaluation,
we use a subset of tasks from the Shopping split of the WebArena benchmark \citep{zhou2023webarena} with automatic evaluators based on URLs.
We leverage our fine-tuned Mistral-7B intent predictors from the Mind2Web experiments without any modifications.
As Mind2Web \citep{mind2web_deng2024} does not include stop actions in its dataset, we perform step-wise evaluations to check task completion for all the compared methods.
We employ the observation processing and element-ranking model described in \Cref{sec:appendix_compared_methods} for all the methods compared in this online evaluation.

\end{document}